# A Novel and Robust Evolution Algorithm for Optimizing Complicated Functions

Gao Yifeng, Gong Shuhong, and Zhao Ge

*Abstract*-In this paper, a novel mutation operator of differential evolution algorithm is proposed. A new algorithm called divergence differential evolution algorithm (DDEA) is developed by combining the new mutation operator with divergence operator and assimilation operator (divergence operator divides population, and, assimilation operator combines population), which can detect multiple solutions and robustness in noisy environment. The new algorithm is applied to optimize Michalewicz Function and to track changing of rain-induced-attenuation process. The results based on DDEA are compared with those based on Differential Evolution Algorithm (DEA). It shows that DDEA algorithm gets better results than DEA does in the same premise. The new algorithm is significant for optimizing and tracking the characteristics of MIMO (Multiple Input Multiple Output) channel at millimeter waves.

## I. INTRODUCTION

Genetic Algorithm (GA) is a new optimization method which is influenced by evolutionary theory given by Darwin. GA is aiming at searching for the optimum solution by simulating biological evolution [1]. GA is unique from other algorithms for better performance in searching for multiple and complex objective functions. Unfortunately, some problems, such as selecting parameters and convergence of the GA etc., cannot be completely solved. So, many improved neotype algorithms are created, for instance, parallel Genetic Algorithm, niche algorithm, hybrid Genetic Algorithm, differential evolution algorithm and so on [3-12], all of which are widely used in applications and performing well in practice. But those of the parallel Genetic Algorithm and differential evolution algorithm can only choose initial population using empirical method. Divergence operators are recommended for choosing initial population, which can enlarge the searching area and finally get the optimum solution. Some novel divergence operators are given in recent literatures, such as chromosomal, speciation, gender divergence, and the group divergence [3-12]. The same feature of these operators is separating evolution by dividing a single population into a lot of subspecies to avoid the elitists hindering the evolution. But how to improve efficiency in cluster algorithm is the caused new problem.

In this paper, a novel mutation operator of differential evolution algorithm is proposed. A new algorithm called divergence differential evolution algorithm (DDEA) is developed by combining the new mutation operator with divergence operator and assimilation operator, which can detect multiple solutions and robustness in noisy environment. The new algorithm is applied to optimize Michalewicz Function and to track changing of rain-induced-attenuation process. The results based on DDEA are compared with those based on Differential Evolution Algorithm (DEA). It shows that DDEA algorithm gets better results than DEA does in the same premise. The new algorithm is significant for optimizing and tracking the characteristics of MIMO channel at millimeter waves.

## II. IMPROVED DIFFERENTIAL EVOLUTION ALGORITHM

Suppose $N$ is the number of individuals, $x_i$ is the $i^{th}$ individual, $x_{j,i}$ is the $j^{th}$ genes located in $i^{th}$ individual, f is the objective function, $g$ indicates $g^{th}$ generation. The traditional DEA can be described as follow 4 steps:

1. Initializing population:
$$x_{j,i} = \xi(b_{j,U} - b_{j,L}) + b_{j,L} \qquad (1)$$

Where, $b_L$ and $b_U$ indicate the lower and upper bounds, $\xi$ indicates a uniformly distributed random number within interval [0, 1).

2. Computing difference vector based mutation
$$v_{j,g} = y_{i,g} + F(x_{r_1,g} - x_{r_2,g}) \qquad (2)$$

Where, $r_1$, $r_2$ are randomly selected integers from (1, $N$). The equation (2) is computed for each $y_{i,g}$ vector, $y_{i,g}$ is $i^{th}$ individual in $g^{th}$ generation, $F$ is a constant between 0 and 1.

3. Crossover
$$u_{i,g} = u_{j,i,g} = \begin{cases} v_{j,i,g} & if(\xi \leq Cr) \\ x_{j,i,g} & otherwise \end{cases} \qquad (3)$$

Where, $Cr$ is a constant in [0 1]. Equation (3) is the most common form of crossover.

4. Selecting

The vector with the lowest objective function value can go into the next generation, according to the simple one-to-one

Manuscript received July 1, 2011. This work has been supported by "The National Natural Science Foundation of China (61001065)".

Gao Yifeng is with the School of Science, Xidian University, Xi'an, China (phone: +86-15029980593; e-mail: gaoyfeng2010@hotmail.com).
Gong Shuhong is with the School of Science, Xidian University, Xi'an, China (e-mail: ljbrp2003@yahoo.com.cn).
Zhao Ge is with the School of Science, Xidian University, Xi'an, China (e-mail: zhaoge123@gmail.com).

survivor selection principle.
Then

$$x_{i,g+1} = \begin{cases} u_{i,g} & \text{if } f(u_{i,g}) \leq f(x_{i,g}) \\ x_{i,g} & \text{otherwise} \end{cases} \quad (4)$$

If $p(x_i)$ indicates the probability density function (pdf) for $i^{th}$ generation, equation (2) can be replaced by

$$v_{i,g} = p(x_i) + F(p(x_i) - p(x_i)) \quad (5)$$

In fact, the geographic condition of the objective function can be reflected by the distribution of population. The distribution of population defined by the objective function is shown in Fig.1.a and Fig.1.b. It can be found that the density near the peak of population is higher than that near the trough.

The approximation of probability density is shown in Fig.1.c, which is got with the help of Gaussian distribution. It can be found that the more nearer to the peak, the more similar it is to the Gaussian distribution.

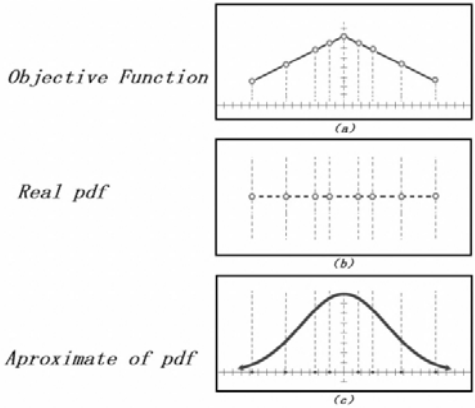

Fig.1: The estimated probability density distribution in the condition that the objective function is symmetry. (a): symmetry objective function. (b): the real probability density function. (c): the estimated probability density function.

The distribution can move nearer to the optimal point under the condition as in Fig.2.a. In Fig.2.b, dummy line denotes the distribution of the 1$^{st}$ generation, dotted line denotes the distribution of 2$^{nd}$ generation, and solid line denotes the final distribution of 3$^{rd}$ generation. Theoretically, the Gaussian distribution can replace the real distribution near the peak while the sample is large enough, according to central limit theorem.

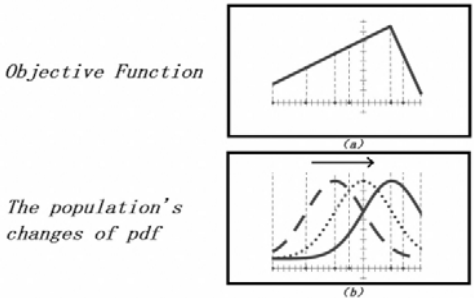

Fig.2: The estimated probability density distribution in the condition that the objective function is asymmetry. (a): symmetry objective function. (b): the population's changes of probability density function.

Assuming that $x$ follows Gaussian density function with $\mu$ and $\sigma$, equation (5) can be written as

$$\begin{aligned} v_{i,g} &= N(\mu, \sigma) + F(N(\mu, \sigma) - N(\mu, \sigma)) \\ &= N(\mu, (1+2F)\sigma) \end{aligned} \quad (6)$$

So, equation (2) can simply be replaced by equation (6). Comparing equation (2) and equation (6), equation (6) has two merits: First, equation (6) only needs one random number, but equation (2) needs three random numbers. So efficiency can be increased in large-scale computing by equation (6). Second, equation (6) can search a continuous range, but equation (2) has only $N^3$ discrete results. So equation (6) gets more diversity for population.

### III. DIVERGENCE AND ASSIMILATION

Gaussian distribution can describe an evolution strategy in theory. But, in most situations, the objective function is complex, containing many local optimal points. In this case, it is unwarranted that the population follows a single Gaussian distribution. The population follows a Gaussian mixture distribution. But, Gaussian mixture distribution doesn't have the good features - addition of complex pdf can be replaced by addition of parameters - which the Gaussian distribution has. In order to solve this problem, a divergence operator is taken into account, which can divide a Gaussian mixture distribution into many Gaussian distributions. Each Gaussian distribution indicates an independent evolution group. This operator can decrease useless searching range. Fig.3 shows the computing process of the divergence operator.

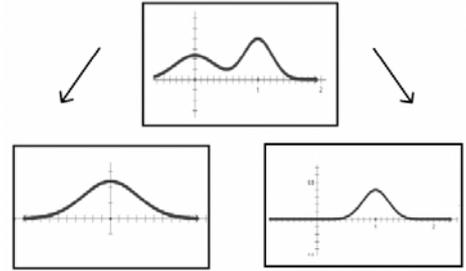

Fig.3: the work of the divergence operator.

Let $|S_i|$ indicates the individual in the range of $\|\mu - x\|_2 \leq l$ of $i^{th}$ generation, define

$$dive_i = \frac{|S_i|}{N} \quad (7)$$

$$k_i = a \frac{l}{\|\max(P_i) - \min(P_i)\|_2} \quad (8)$$

Where, $P_i$ indicates all individuals of the population of $i^{th}$ generation, $a$ is a parameter computed by:

$$a = \frac{S_1}{2k_1} \quad (9)$$

If $dive_i < k_i$, the population still follows Gaussian distribution; otherwise, the population should be separated into two clusters using cluster algorithm. The quantity of individuals in a cluster should be more than $N/10$ for protecting reliability of the cluster. If the cluster is meaningless, the divergence operator interrupts.

On the other hand, the overwhelm groups will create too many individuals, which will cost too much resource. Therefore, it is necessary to use an operator to delete the useless groups. So assimilation operator is needed. Fig.4 shows the process of the assimilation operator. The two groups combine into one because they are too much similar.

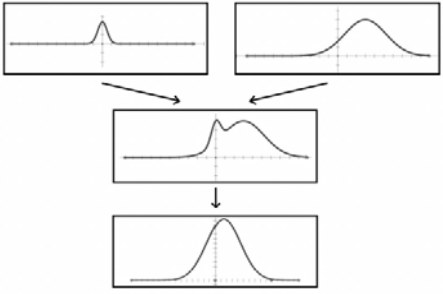

Fig. 4: the work of assimilation operator.

Suppose, $\varepsilon = \|\mu_1 - \mu_2\|_2$, $a'$ is a constant, if $\varepsilon \leq a'$, then

$$\overline{\mu} = \frac{\mu_1 + \mu_2}{2}, \overline{\sigma} = \frac{\sigma_1 + \sigma_2}{2} \quad (10)$$

Fig.5 shows the process of the combination of both operators mentioned above. In the first intermediate process, the density function of population becomes multimodal for the first time, which divides into two groups with separate evolution. In the second intermediate process, the first group became possible to be multimodal again, the operator works like the first one. The second group keeps evolving without division. In the 3$^{rd}$ intermediate process, the 2$^{nd}$ group and 3$^{rd}$ group combine into a single one.

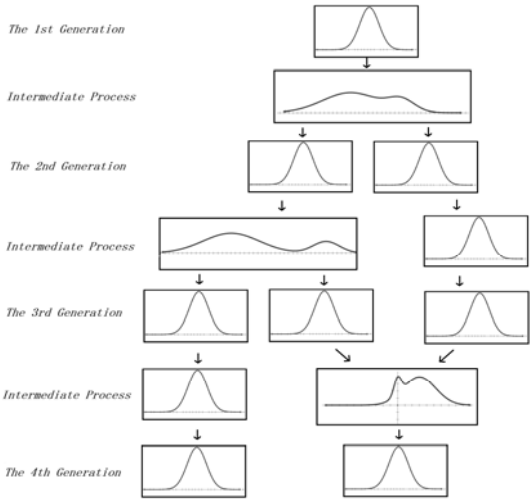

Fig. 5: The co-work of differential operator and assimilation operator.

Assimilation operator and divergence operator can turn the difference evolution algorithm into parallel algorithm, expanding the ability to search more complex objective functions without sacrificing efficiency, and, the errors made by the subjective selection can be erased.

## IV. THE ANALYSIZE OF TEST RESULTS

In this section, the effect of the DDEA is tested by optimizing Michalewicz function and tracing dynamic rain-induced attenuation time series.

(1) Searching Michalewicz Function

Michalewicz is the optimal goal. The function is described as [11]:

$$f(x, y) = -\sin(x)(\frac{\sin x^2}{\pi})^{20} + \sin(y)(\frac{\sin y^2}{\pi})^{20} \quad (11)$$

The graph of equation (11) is shown in Fig. 6.

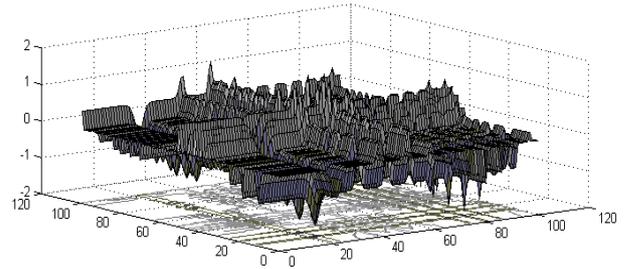

Fig. 6: Michalewicz function

The parameters selected for the DDEA and DEA are listed in Table 1.

|  | DEA | DDEA |
| --- | --- | --- |
| Initial Population | 100 | 100 |
| $F$-Value | 0.8 | 0.8 |
| Initial Low Band | -1 | -1 |
| Initial Up Band | 1 | 1 |
| $l$-Value |  | 1 |
| $a'$-Value |  | 0.1 |

Tab. 1: The Initial parameters of DEA and DDEA

Fig. 7 shows the difference between the result of DDEA and DEA. The calculating times for individuals are the same both in DDEA and DEA. For example: there are four generations, the group numbers within each generation are 1, 2, 4 and 8 in DDEA, and each group contains 100 individuals. (1+2+4+8)*100=1500, thus DDEA and DEA both need 1500 times calculation.

It can be concluded from Fig. 7 that the optimal point worked out by DDEA is better than the one worked out by DEA in term of stability.

(2) Tracing dynamic rain-induced attenuation time series

The database is selected from Rain-Attenuation Series of Xi'an Shanxi Province on July 29th, 2010. The latter series are to be calculated by DDEA and DEA with earlier series, and deviation between prognostic results and measured results is set for our objective function.

Due to the highly random objective series, the algorithm is more likely to be disturbed. Therefore, it is an appropriate tool to test the stability of DDEA.

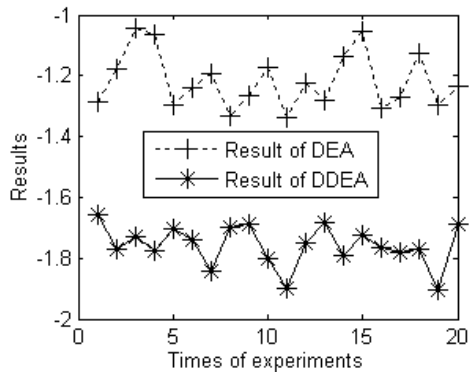

Fig. 7: The result of DEA and DDEA of Michalewicz function.

Fig. 8 shows the prognostic series by both DDEA and DEA in coordinate of logarithm. Fig.8 shows that better accuracy and stability appear in DDEA than in DEA.

V. CONCLUSION

A new evolution algorithm based on divergence evolution algorithm is proposed. It has well performs in optimizing Michalewicz function and tracing dynamic rain-induced attenuation time series. The new algorithm is significant for optimizing and tracking the characteristics of MIMO channel at millimeter waves. For future research, the new type of divergence evolution of mutation operators as described in this paper can be used to design better algorithm.

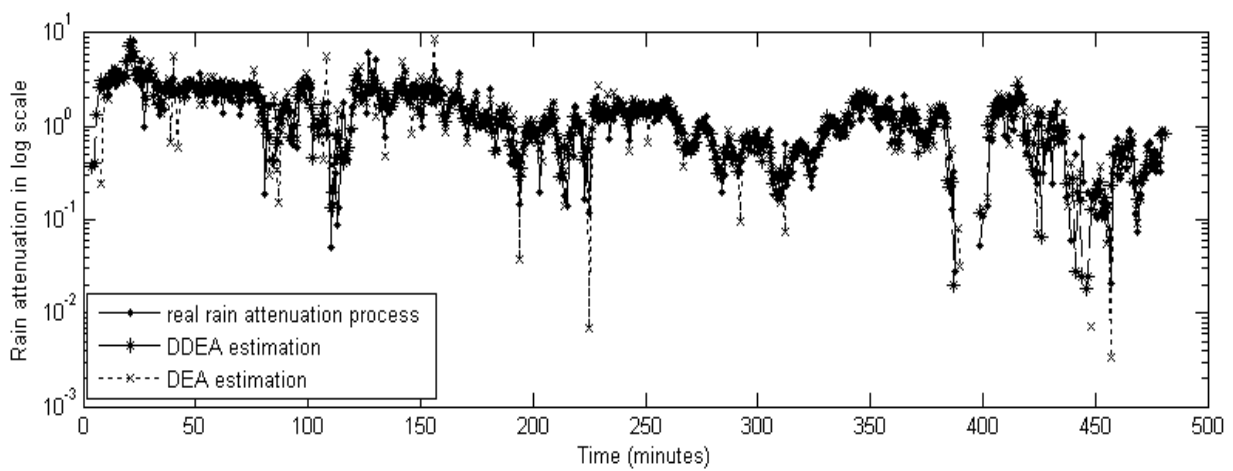

Fig. 8: The estimation of rain-attenuation process using DEA and DDEA.